\documentclass[conference]{IEEEtran}
\IEEEoverridecommandlockouts
\usepackage{cite}
\usepackage{amsmath,amssymb,amsfonts}
\usepackage{algorithmic}
\usepackage{graphicx}
\usepackage{textcomp}
\usepackage{xcolor}
\usepackage{hyperref}
\usepackage{array, cellspace}

\def\BibTeX{{\rm B\kern-.05em{\sc i\kern-.025em b}\kern-.08em
    T\kern-.1667em\lower.7ex\hbox{E}\kern-.125emX}}
\begin{document}

\title{Improving Extrinsics between RADAR and LIDAR using Learning
}

\author{\IEEEauthorblockN{Peng Jiang}
\IEEEauthorblockA{\textit{J. Mike Walker '66 Dept. of Mechanical Engineering} \\
\textit{Texas A\&M University}\\
College Station, USA \\
maskjp@tamu.edu}
\and
\IEEEauthorblockN{Srikanth Saripalli}
\IEEEauthorblockA{\textit{J. Mike Walker '66 Dept. of Mechanical Engineering} \\
\textit{Texas A\&M University}\\
College Station, USA \\
ssaripalli@tamu.edu}
}

\maketitle

\begin{abstract}
LIDAR and RADAR are two commonly used sensors in autonomous driving systems. The extrinsic calibration between the two is crucial for effective sensor fusion. The challenge arises due to the low accuracy and sparse information in RADAR measurements. This paper presents a novel solution for 3D RADAR-LIDAR calibration in autonomous systems. The method employs simple targets to generate data, including correspondence registration and a one-step optimization algorithm. The optimization aims to minimize the reprojection error while utilizing a small multi-layer perception (MLP) to perform regression on the return energy of the sensor around the targets. The proposed approach uses a deep learning framework such as PyTorch and can be optimized through gradient descent. The experiment uses a 360-degree Ouster-128 LIDAR and a 360-degree Navtech RADAR, providing raw measurements. The results validate the effectiveness of the proposed method in achieving improved estimates of extrinsic calibration parameters.
\end{abstract}


\section{Introduction}
With the advancement of autonomous vehicle technology, there is an increasing need for effective solutions in adverse weather conditions, such as rain and snow. While CAMERA and LIDAR sensors have proven successful under normal conditions, inclement weather can significantly affect their performance. RADAR sensors, like those produced by Navtech \cite{barnes_oxford_2020}, offer a promising alternative. With a longer wavelength, RADAR sensors are less impacted by small particles like dust, fog, rain, or snow that can impair the performance of CAMERA and LIDAR. Additionally, RADAR has a longer detection range and the ability to penetrate materials, enabling the detection of objects that are beyond the line of sight of LIDAR sensors. These features make RADAR ideal for use in adverse weather conditions.

The publication of the Oxford RADAR RobotCar Dataset \cite{barnes_oxford_2020} in 2020 has provided a valuable resource for researchers, containing a rich collection of LIDAR, CAMERA, and RADAR data. This dataset has been instrumental in advancing autonomous vehicle research and has led to notable progress in fields such as odometry \cite{barnes_oxford_2020,burnett_radar_2021,barnes_under_2020,burnett_we_2021}, place recognition \cite{kim_scan_2022,kim_scan_2018}, and semantic segmentation \cite{ouaknine_multi-view_2021,prophet_semantic_2019,rebut_raw_2022,kaul_rss-net_2020}. Other datasets like \cite{zhou_towards_2022,burnett_boreas_2022,ouaknine_carrada_2020,sengupta_automatic_2022} have also been made accessible to researchers, further fueling the development of RADAR research. We expect RADAR and its integration with other sensors to become a growing area of research and application in the future, with great advancements driven by the availability of large amounts of public data.

While RADAR sensors have the potential to provide robust performance in inclement weather, they also have a coarser spatial resolution and higher noise compared to LIDAR sensors, making the task of calibrating them challenging. Extrinsic calibration between RADAR and other sensors has received relatively less attention compared to the calibration between LIDAR and CAMERA sensors \cite{zhou_towards_2022,wang_soic_2020,jiang_semcal_2021,lv_lccnet_2021,zhao_calibdnn_2021}.

The extrinsic 6-DoF calibration between LIDAR and RADAR sensors can be difficult due to the limited resolution and information loss in the third dimension of RADAR measurements. Previous methods, such as \cite{persic_extrinsic_2019}, utilized curve models to model the relationship between elevation and RADAR cross-section (RCS). However, these models have limitations as they only consider a subset of data and do not account for various factors that impact the RADAR return signal, such as shape, material, direction, and the RADAR's power and frequency.
Recent advancements in neural networks, such as the Neural Radiance Field (NeRF) model \cite{mildenhall_nerf_2020}, demonstrate the ability to model complex relationships between sensors, poses, and scenes, including scene geometry. These models can learn to represent a scene using images from various views and estimate the pose during the training process \cite{lin_barf_2021}.
In this paper, we present a method to calibrate LIDAR and RADAR by using a small Multi-Layer Perceptron (MLP) to model the target as a collection of RADAR return energy. The model is trained to learn the correlation between the RADAR sensor pose relative to the target and the return energy, allowing it to be used for calibration. This is treated as a regression problem. Along with the regression loss, we also use reprojection loss and a ray pass loss to achieve calibration. The details of these losses will be described in \ref{ssec:optimization}.
\section{Related works}
\subsection{Multisensor Calibration}
In the calibration of RADAR and other sensors, two primary challenges exist: (1) recovering the missing third-dimension information from RADAR and other sensors, and (2) ensuring that reflectors are visible to all sensors. Traditional methods have addressed these issues in various ways. For example, \cite{el_natour_radar_2015} proposed a 3D reconstruction method based on sensor geometry and a calibration facility using a Luneburg lens and differently-colored corner reflectors. \cite{persic_extrinsic_2017} designed a compact target and proposed two-step optimizations for 6-DOF calibration between RADAR and LIDAR. Subsequently, \cite{persic_extrinsic_2019} added chequerboard patterns to enable calibration between RADAR, LIDAR, and CAMERA. \cite{domhof_extrinsic_2019} utilized a styrofoam board with four circular holes and a corner reflector for sparse LIDAR beam detection. More recent tools like \emph{radar\_to\_lidar\_calib} \cite{burnett_radar_to_lidar_calib_2023} and \emph{OpenCalib} \cite{yan_opencalib_2022} have also emerged, offering calibration capabilities for multiple sensor types without targets. Our novel approach distinguishes itself from these existing methods by utilizing a more comprehensive representation of the target, rather than only considering data with the strongest return energy. We achieve this by training a small neural network to represent the target, which allows our method to account for both the detected target center, characterized by high return energy, and the data from the surrounding neighborhood. This innovative approach leads to a more accurate and nuanced calibration process, ensuring enhanced performance in multi-sensor systems.
\begin{center} 
\begin{figure}[t!]
\begin{tabular}{c c} 
\includegraphics[width=0.22\textwidth]{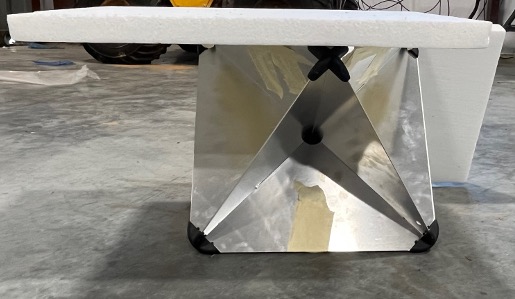}&\includegraphics[width=0.22\textwidth]{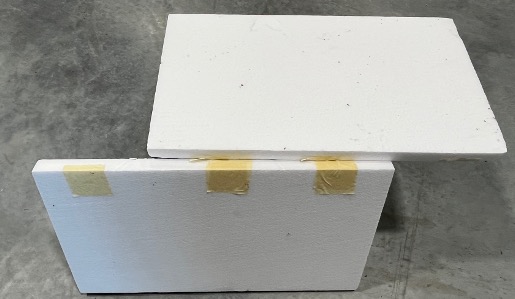}\\
{\scriptsize (a) Side view} & {\scriptsize (b) Front view}\\
\end{tabular} 
\caption{Aluminum cctahedral RADAR reflector front with two side covered by two polystyrene foams} 
\label{fig:reflector}
\end{figure}
\end{center} 
\subsection{RADAR and Neural Network}
Deep learning has demonstrated its effectiveness in various applications, including autonomous driving. While CAMERA and LIDAR have benefited from extensive research, the application of deep learning to RADAR remains under-explored, as evidenced by the limited literature \cite{zhou_towards_2022}. Notable works on CAMERA and LIDAR calibration using deep learning, such as \cite{wang_soic_2020,jiang_semcal_2021,lv_lccnet_2021,zhao_calibdnn_2021}, have employed supervised or unsupervised approaches, but similar studies for RADAR and other sensor calibrations are still lacking. These methods leverage neural networks to identify correlations between modalities and learn common features, either guided by ground truth or an unsupervised loss. A promising future direction could involve unsupervised metric feature learning techniques, such as \cite{burnett_radar_2021,barnes_under_2020}, for discovering shared features between RADAR and other modalities, as demonstrated in \cite{jiang_contrastive_2022}. Implicit neural representations have recently gained popularity in representing scenes for images and 3D point clouds. Gradient descent allows neural networks to learn color and geometry information from image data \cite{mildenhall_nerf_2020} or point cloud data \cite{pan_voxfield_2022}, with or without pose information. However, to the best of our knowledge, this approach has not yet been applied to RADAR data. Our work stands out by exploring the untapped potential of deep learning in RADAR calibration, addressing the gap in existing research and offering a novel contribution to the field.
\section{Method}
\subsection{Calibration Target Design}
Calibrating both LIDAR and RADAR requires a target that is visible to both sensors. For RADAR, it is desirable to have a target with a high RADAR Cross-Section (RCS) to improve the detection rates. Marine aluminum RADAR reflectors (see Fig. \ref{fig:reflector} (a)), which are cheap and readily available, can be used for this purpose, but they can affect the LIDAR measurements. Furthermore, their small size makes them difficult to detect by LIDAR. To overcome this, we use two polystyrene foams to cover two sides of the reflector and construct a non-symmetrical shape as see Fig. \ref{fig:reflector} (b), which has a larger visible area for LIDAR, making it easier to register and determine the reflector's position in the scene, which will be used for calibration.
\begin{figure}[t!]
   \centering
    \includegraphics[width=0.45\textwidth]{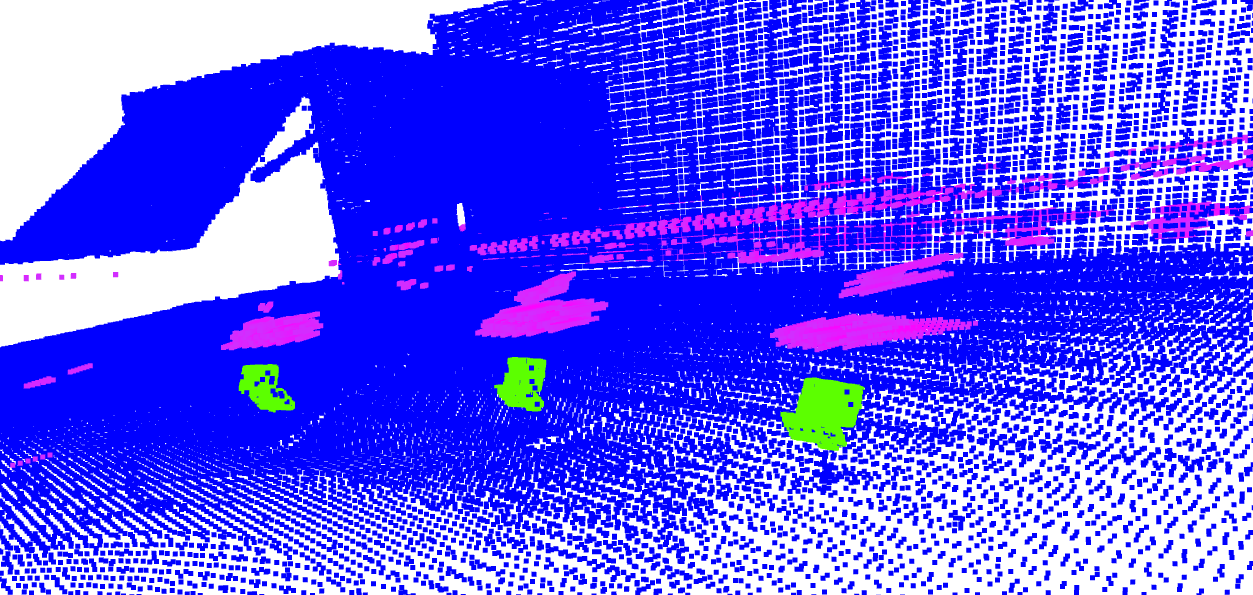}
    \caption{Registered LIDAR Point Cloud (blue), Filtered RADAR measurements(Purple), Extracted Target Point Cloud(green)}
    \label{fig:register}
\end{figure}
\subsection{Data Collection}\label{ssec:data_collection}
To guarantee accurate and undistorted data from both LIDAR and RADAR sensors, we leverage velocity information obtained from the Inertial Measurement Unit (IMU) sensor to identify static frames in LIDAR and RADAR data. We used the FPFH feature \cite{rusu_fast_2009} and RANSAC global registration to register all the static LIDAR frames together, resulting in an initial pose of all the LIDAR data in the scene. We then refined the pose of each LIDAR frame using the Iterative Closest Point (ICP) algorithm to generate a dense LIDAR point cloud, which was then utilized for target detection to prevent any incomplete frames. By collecting data from multiple robot movements, we obtained each LIDAR frame’s position within the registered point cloud, thus eliminating the need for target extraction for each frame.  For the RADAR target extraction, we first filtered out most of the raw RADAR by setting a threshold. Then we used raw LIDAR and RADAR transformation to get the raw data target measurement, which was obtained by manually measuring. We defined the target center as the first measurement in the RADAR data. The data collection and pre-processing results  is shown in Fig.\ref{fig:register}.
\begin{figure}[t!]
   \centering
    \includegraphics[width=0.5\textwidth]{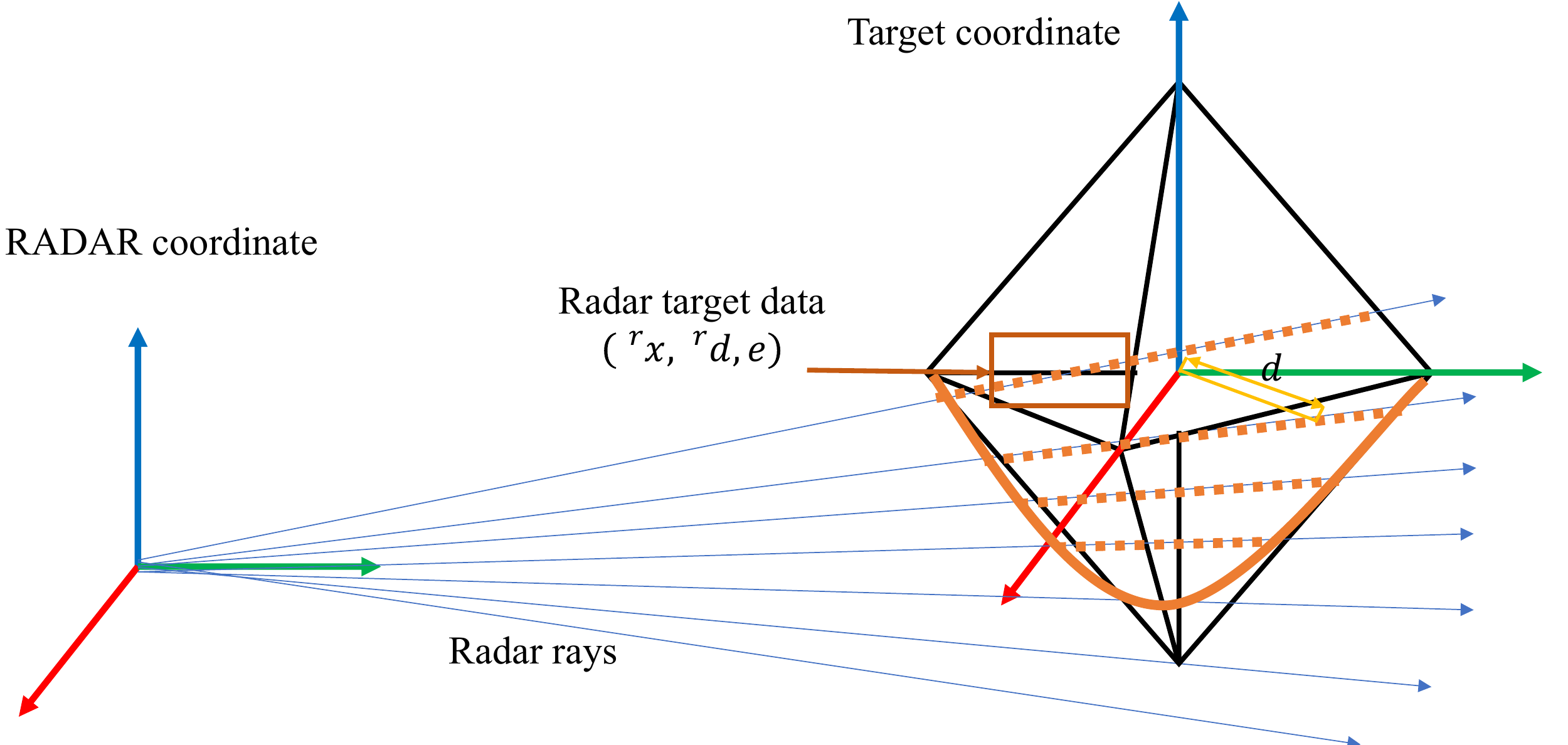}
    \caption{Sample the RADAR measurement data $\left({}^r\boldsymbol{x}_i,{}^r\boldsymbol{d}_i,\boldsymbol{e}_i\right)$ around the target centers; Compute the data distance $d$ between the LIDAR target center and the RADAR target center}
    \label{fig:nerf}
\end{figure}
\subsection{Optimization}\label{ssec:optimization}
We develop our calibration method using a gradient descent optimization framework implemented with PyTorch. Our objectives for the calibration are three-fold: 1) minimizing the reprojection error, 2) reducing the regression error of an MLP, and 3) minimizing the ray pass loss, which considers whether a ray passes through the target or not. The transformation matrix $\boldsymbol{T}$ between the LIDAR and RADAR sensors can be represented in many ways. While optimizing, the rotation matrix $R(\boldsymbol{w})$ was represented as axis-angle representation $\boldsymbol{w}$ ($\mathfrak{so}(3) \rightarrow \operatorname{SO}(3)$). While evaluation, we represent the rotation using Euler angles $\boldsymbol{\theta}=\left[\theta_x,\theta_y,\theta_z\right]$. The translation is represented by three variables $\boldsymbol{t}=\left[t_x,t_y,t_z\right]$.
\subsubsection{Reprojection Loss}
To compute reprojection loss, we first convert the LIDAR target center point ${}^l\boldsymbol{x}_{l, i}$ from the LIDAR sensor coordinate to the RADAR sensor coordinate using the following equation:
\begin{equation}
{ }^r \boldsymbol{x}_{l, i}={ }^r_l R \cdot{ }^l \boldsymbol{x}_{l, i}+{ }^r_l\boldsymbol{t}
\end{equation}
Where ${ }^r_l R$ is the rotation matrix and ${ }^r_l\boldsymbol{t}$ is the translation vector. Next, the point ${}^r\boldsymbol{x}_{l, i}$ is converted to spherical coordinate ${}^r\boldsymbol{s}_{l, i} = [{}^r r_{l, i},{}^r\theta_{l, i},{}^rz_{l, i}]$.
The RADAR measurement ${}^r\boldsymbol{s}_{r, i} = [{}^rr_{r, i},{}^r\theta_{r, i}, -]$ is assumed to be the ground truth. The absolute differences loss ($L_1$) is applied to the azimuth angle and radial distance between the transformed LIDAR target center data and the RADAR measurement.
\begin{equation}
l_{rep}=w_{r}L_1({ }^r r_{r},{ }^r r_{l})+w_{\theta}L_1({ }^r \theta_{r},{ }^r \theta_{l})
\end{equation}
where $w_{\theta}$ and $w_{r}$ are two weights that can be set manually to adjust the optimization direction and balance the relative importance of radial distance and azimuth angle, due to their differing value scales.
\begin{figure}[t!]
\begin{tabular}{c c} 
\includegraphics[width=0.22\textwidth]{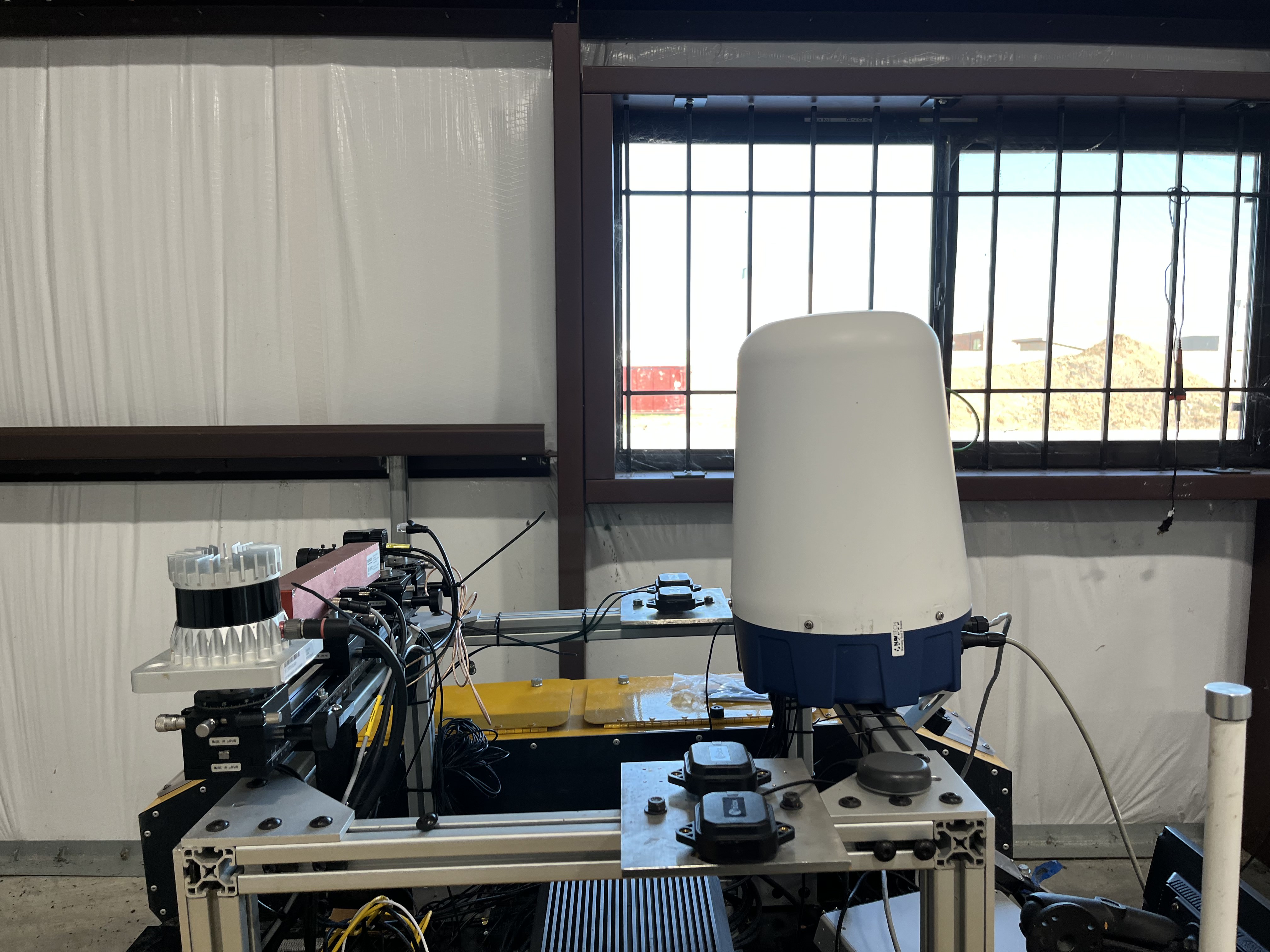}&\includegraphics[width=0.22\textwidth]{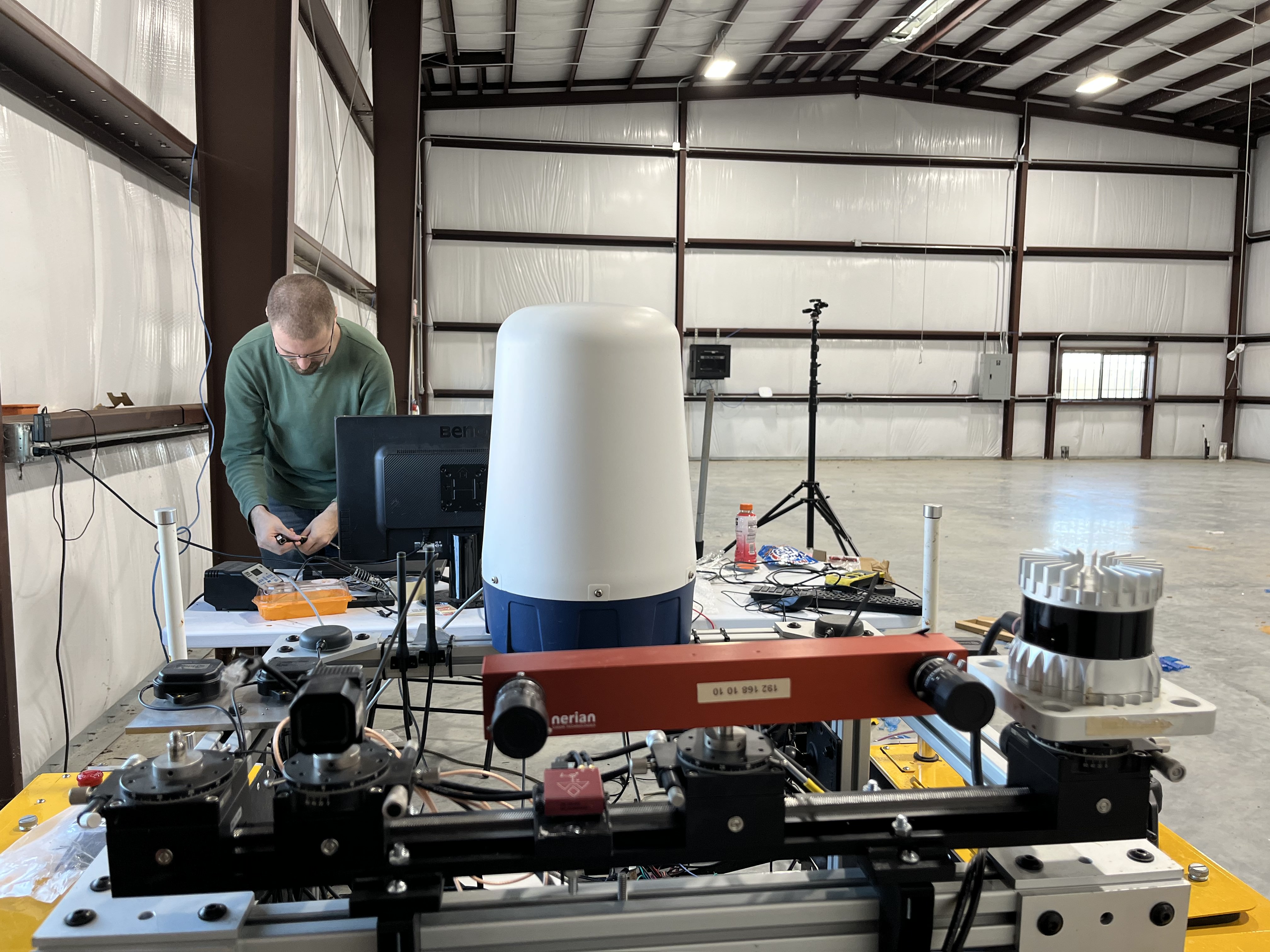}\\
{\scriptsize(a) Side view} & {\scriptsize(b) Front view}\\
\end{tabular} 
\caption{Clearpath Warthog with one Ouster OS1 128 Channel LIDAR and one Navtech CIR-DEV RADAR} 
\label{fig:sensors}
\end{figure}

\subsubsection{Regression Loss}
Getting a 6 DoF calibration between LIDAR and RADAR is challenging due to the missing 3D information in RADAR measurement. However, as shown in \cite{persic_extrinsic_2019}, the RCS is correlated with the target's 3D position in the RADAR coordinate, leading to the modeling of the relationship between RCS and target position as a curve model in the paper. Inspired by this work and recent advancements in NeRF, we use a small Multi-Layer Perceptron (MLP) to model the relationship between the RADAR sensor pose relative to the target and the return energy. Rather than solely relying on the target center to model the relationship, we also incorporate the neighborhood data surrounding the target centers for a more comprehensive representation. See. Fig.\ref{fig:nerf}, we sample the RADAR measurement data around the target centers and represent the data as $\left({}^r\boldsymbol{x}_{r,i},{}^r\boldsymbol{d}_{r,i},e_i\right)$, where ${}^r\boldsymbol{x}_{r,i}$ denote the position of the RADAR data in RADAR coordinate, ${}^r\boldsymbol{d}_{r,i}$ represent the ray direction from the RADAR sensor origin to the measurement, and $e_i$ represent the return energy, which is raw data value of the Navtech RADAR.

In the optimization process, we transform the RADAR measurement $\left({}^r\boldsymbol{x}_{r,i},{}^r\boldsymbol{d}_{r,i},e_i\right)$ from the RADAR coordinate to the local target coordinate based on the current estimated RADAR-LIDAR extrinsic calibration ${}^w_rT$ and the target pose ${}^w_tT$ in the scene which can be obtained during the data collection steps described in \ref{ssec:data_collection}. After the transformation ,we got data $\left({}^t\boldsymbol{x}_{r,i},{}^t\boldsymbol{d}_{r,i},e_i\right)$ in the target local coordinate.
Based on \cite{mildenhall_nerf_2020} and experiment, we use Eq.\ref{eq:pe} to apply position encoding to each of the three coordinate values in the direction ${}^t\boldsymbol{d}_{r,i}$ and position ${}^t\boldsymbol{x}_{r,i}$ separately. Eq.\ref{eq:pe} is a mapping from $\mathbb{R}$ into a higher dimensional space $\mathbb{R}^{2L}$. Positional encoding facilitates the network to optimize parameters by easily mapping input to higher-dimensional space. \cite{mildenhall_nerf_2020} showed that using a high-frequency function for mapping original input enables better fitting of data that contains high-frequency variation.
\begin{equation}
\begin{aligned}
P(x, 2i) & =\sin \left(2^i\pi x\right) \\
P(x, 2i+1) & =\cos \left(2^i\pi x\right) \\
\end{aligned}
\label{eq:pe}
\end{equation}
We use the encoded direction $P({}^t\boldsymbol{d}_{r,i})$ and position $P({}^t\boldsymbol{x}_{r,i})$ as the input of the MLP and predict the corresponding return energy $\boldsymbol{\hat{e}}_i$.
\begin{equation}
\hat{e}_i=MLP(P({}^t\boldsymbol{x}_{r,i}),P({}^t\boldsymbol{d}_{r,i}))
\end{equation}
During optimization, we minimize the absolute difference ($L_1$ loss) between the ground truth value $\boldsymbol{e_i}$ and the predictions $\boldsymbol{\hat{e}}_i$
\begin{equation}
l_{mlp}=L_1(\hat{e}_i,e_i)
\end{equation}

\subsubsection{Ray-pass Loss}
To ensure that our optimization results align with the properties of the sensors, we added a ray pass loss to optimization objectives. The high return energy of RADAR target measurement should be the result of a ray passing through the target from the RADAR sensor origin. To simplify, we modeled the target as a ball rather than an octahedron. The loss is defined as in Equation \ref{eq:ray}, where $d$ is the distance between the ray and the center of the target in the target coordinate system, and $r$ is the radius of the circumscribed sphere. If the distance is greater than the radius, it indicates that the ray does not pass through the target, and the difference between the distance and radius is used as the penalty. If the distance is smaller than the radius, the penalty is set to zero.
\begin{equation}
    l_{ray} = \max\left(0,d_i-r\right)
\label{eq:ray}
\end{equation}
\subsubsection{Total Losses}
The overall optimization objective is:
\begin{equation}
l=w_{rep}l_{rep}+w_{mlp}l_{mlp}+w_{ray}l_{ray}
\end{equation}
where the $w_{rep}$, $w_{mlp}$ and $w_{ray}$ are weights for each loss. 

\section{Experiments}
\bgroup
\setlength{\tabcolsep}{0pt} 
\renewcommand{\arraystretch}{1} 
\small
\begin{center} 
\begin{figure}[t!]
\begin{tabular}{c c} 
 \multicolumn{2}{c}{\includegraphics[width=0.49\textwidth]{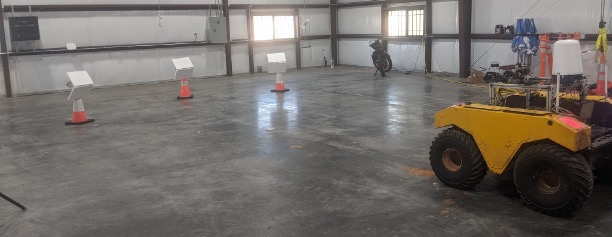}}\\
 \multicolumn{2}{c}{{\scriptsize(a) Sensors and Targets Setup for Data Collection}}
\\
\includegraphics[width=0.245\textwidth]{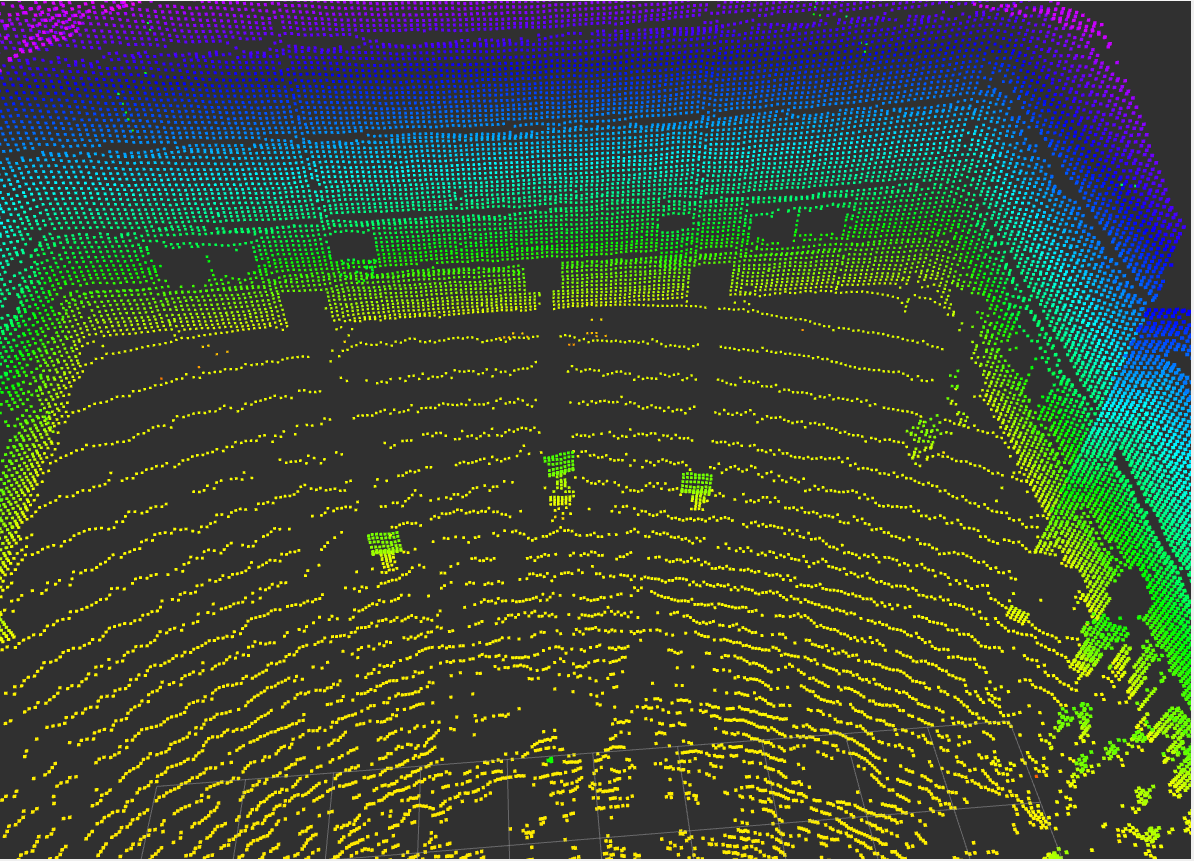}&\includegraphics[width=0.245\textwidth]{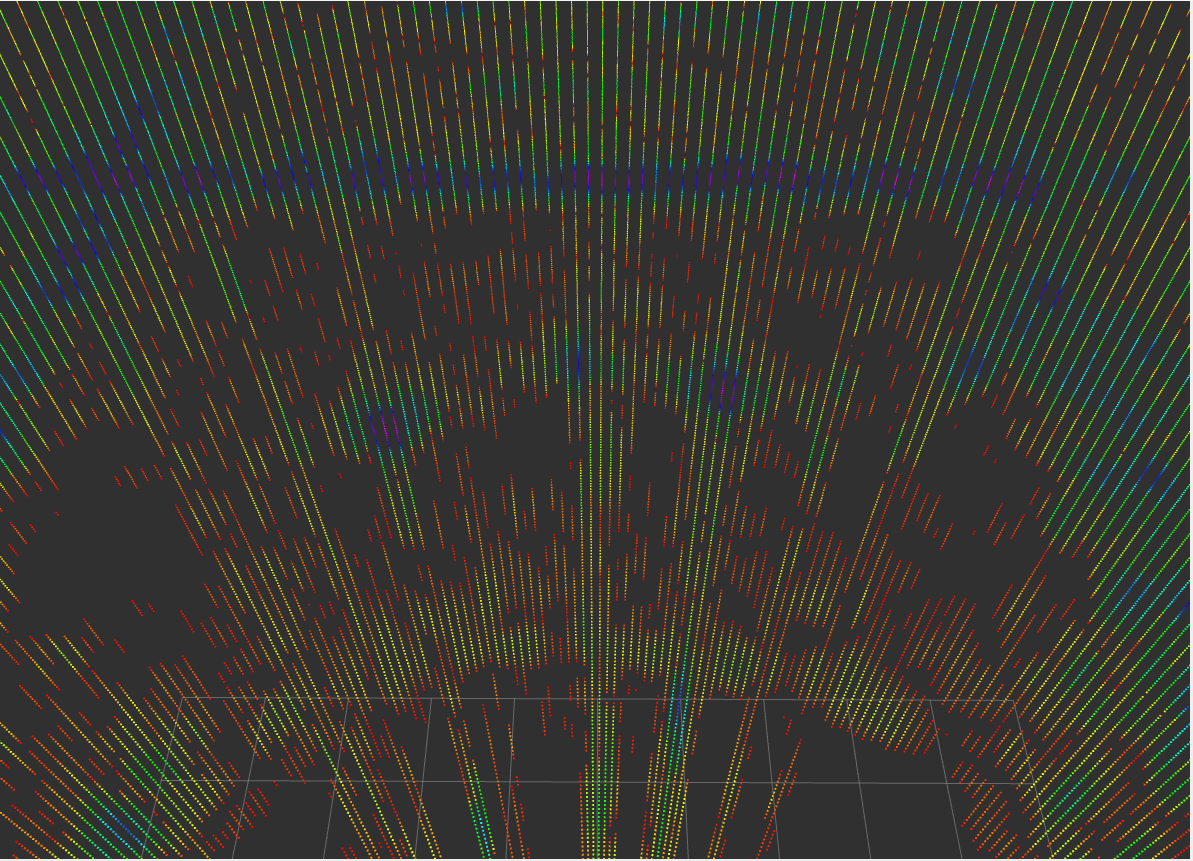}\\
{\scriptsize (b) Raw LIDAR data} & {\scriptsize(c) Raw RADAR data}\\
\end{tabular} 
\caption{Experiment settings: Three targets with fixed positions form a scene, and the robot drives around the targets.} 
\label{fig:experiment}
\end{figure}
\end{center} 
\egroup
\subsection{Sensors and Experiment Setting}
The experiment involved a Clearpath Warthog mobile robot equipped with an Ouster OS1 3D LIDAR and a Navtech CIR-DE RADAR, as depicted in Fig. \ref{fig:experiment}. The LIDAR has 128 channels and 2048 points per channel, with an  45 degrees vertical field of view (FoV) and operates at 10 Hz. The RADAR has a range resolution of 0.044m and azimuth resolution of 0.9 degrees, with an  1.8 degrees vertical field of view (FoV) and works a 4 Hz. Both sensors have  360 degrees horizontal field of view (FoV). The sensor setup is shown in Figure \ref{fig:sensors}. A 3D-printed wedge was attached to the bottom of the RADAR with an incline of around three degrees.

The experiment used three targets, and the robot was positioned at different locations to collect data, as shown in Fig. \ref{fig:experiment}. The experiment was conducted outdoors, and five different target settings were used. After the data processing steps described in \ref{ssec:data_collection}, 104 paired RADAR-LIDAR data were collected. 
\begin{figure}[t!]
   \centering
    \includegraphics[width=0.5\textwidth]{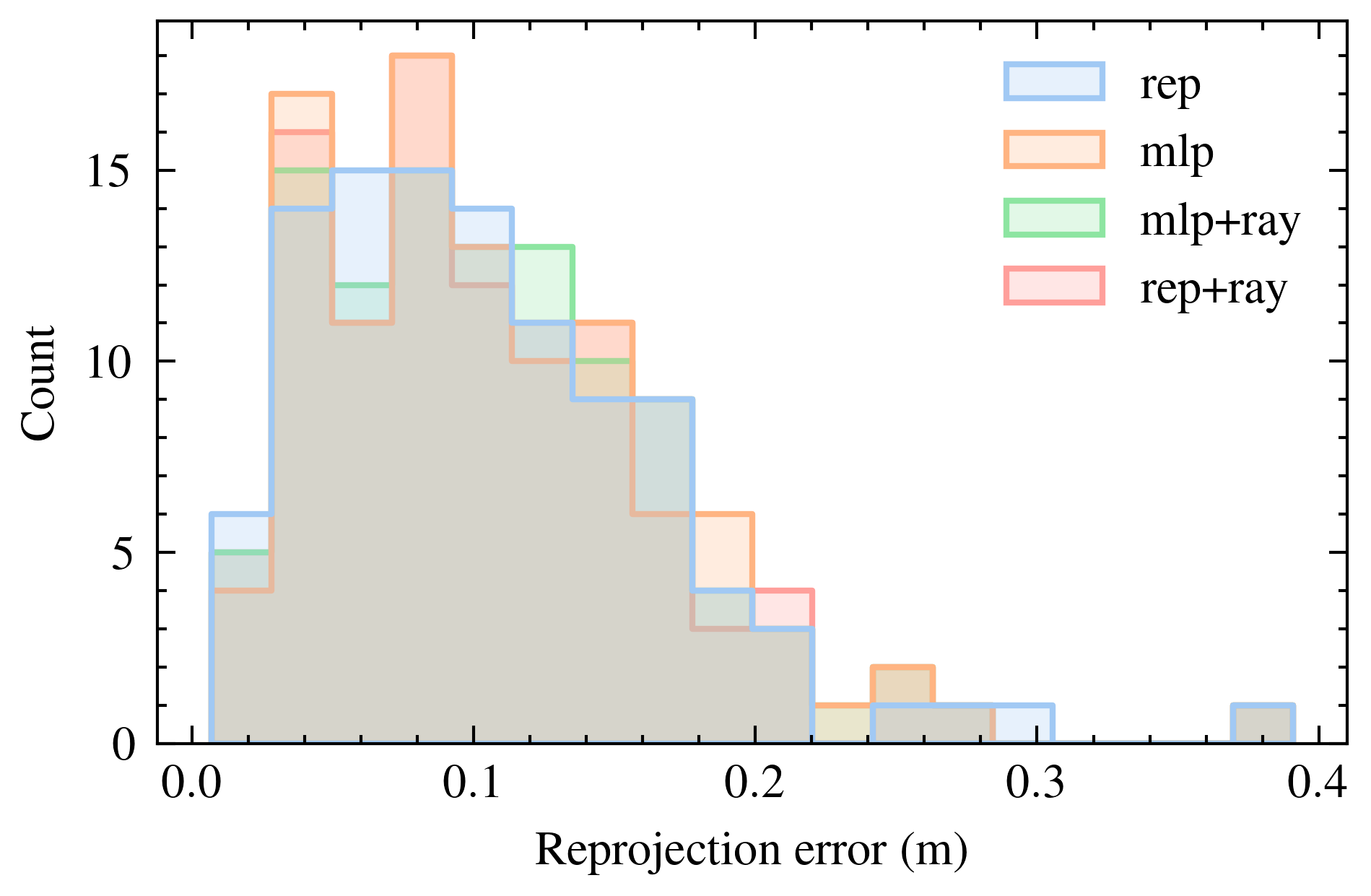}
    \caption{Histogram of reprojection errors for RADAR-LIDAR calibration using different loss configuration}
    \label{fig:reproject_error}
\end{figure}

\subsection{Implementation}
Our extrinsic calibration algorithm was implemented using the PyTorch framework and optimized using the Adam optimizer \cite{paszke_pytorch_2019}. The MLP consisted of four linear layers, each with a hidden size of 128 and using the ReLU activation function. A learning rate of 0.005 was applied to the network, while the learning rates for the rotation and translation parameters were set to 0.005 and 0.001, respectively. During the experiment, the weights for the reprojection error, regression, and ray-pass losses were set to 1000, 1000, and 100, respectively. To compute the regression loss, we sampled data around the estimated RADAR target center within a radius of 0.6 meters.
\begin{table}[b]
\renewcommand{\arraystretch}{1.3}
\caption{RADAR-LIDAR Calibration Results}
\label{table:res}
\centering
\begin{tabular}{c|c|c|c|c|c|c}
\hline
\bfseries Loss &  $\theta_x$ &  $\theta_y$ &  $\theta_z$ & $t_x$ & $t_y$ & $t_z$\\
\hline\hline
initial & 0.00 & 0.00 & 0.00 & 0.50 & -0.25 & 0.05\\
rep & 0.97 & 6.97 & 1.17 & 0.57 & -0.26 & 0.04\\
mlp & 0.26 & 2.04 & 1.01 & 0.57 & -0.25 & 0.05\\
mlp+ray & -0.14 & 2.63 & 1.05 & 0.56 & -0.26 & -0.03\\
rep+ray & 0.08 & 3.35 & 1.07 & 0.57 & -0.26 & 0.01\\
\hline
\end{tabular}
\end{table}
\subsection{Results}
To evaluate the quality of the calibration results, we conducted experiments on a real-world dataset with different loss function configurations: including only the reprojection error loss (\emph{rep}), the reprojection error loss and regression loss (\emph{mlp}), the reprojection error loss, regression loss, and ray-pass loss (\emph{mlp+ray}) and the reprojection error loss and ray-pass loss (\emph{rep+ray}). Table \ref{table:res} presents the initial calibration parameters obtained manually and the results of calibrations using different optimization objectives. The results demonstrate differences, particularly in $\theta_x$ and $\theta_y$, and $t_z$, due to the limitations of the RADAR measurement. However, as previously mentioned, we intentionally introduced a three-degree wedge while installing RADAR sensors. As shown in Table \ref{table:res}, all calibration results reflect this. We used the reprojection error in the RADAR coordinate as the evaluation metric, and Fig. \ref{fig:reproject_error} shows the distribution of the reprojection error. Despite the differences, all four settings have similar reprojection error distributions, which does not necessarily indicate which calibration results are better.
\begin{figure}[t]
   \centering
    \includegraphics[width=0.5\textwidth]{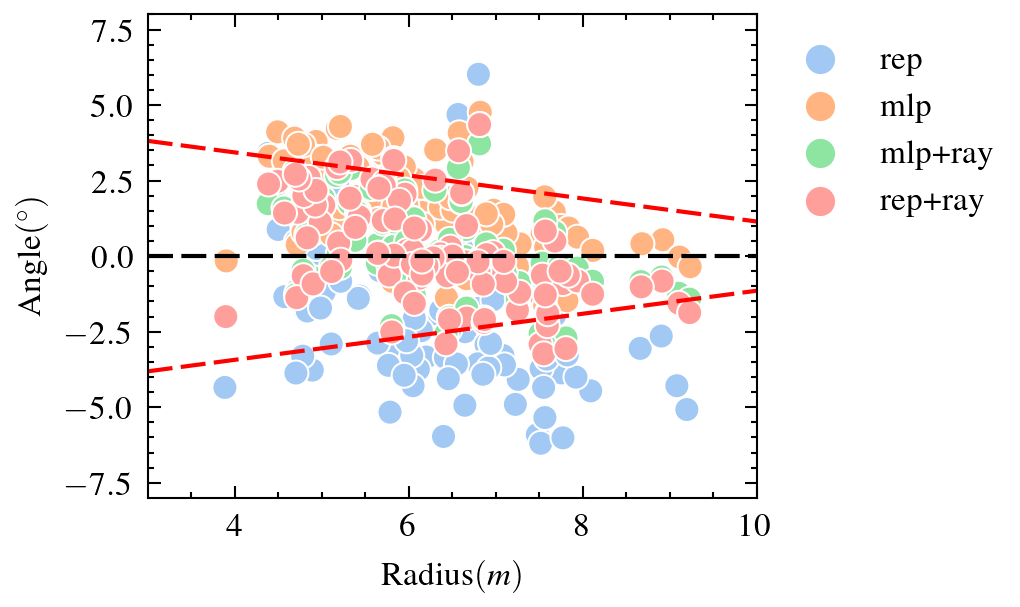}
    \caption{LIDAR target centers distribution in the RADAR coordinate. The x-axis represents the radial distance between the target center and the RADAR origin, the y-axis represents the angle between the RADAR x-y plane and the target centers.}
    \label{fig:theta}
\end{figure}

\begin{figure}[t]
   \centering
    \includegraphics[width=0.5\textwidth]{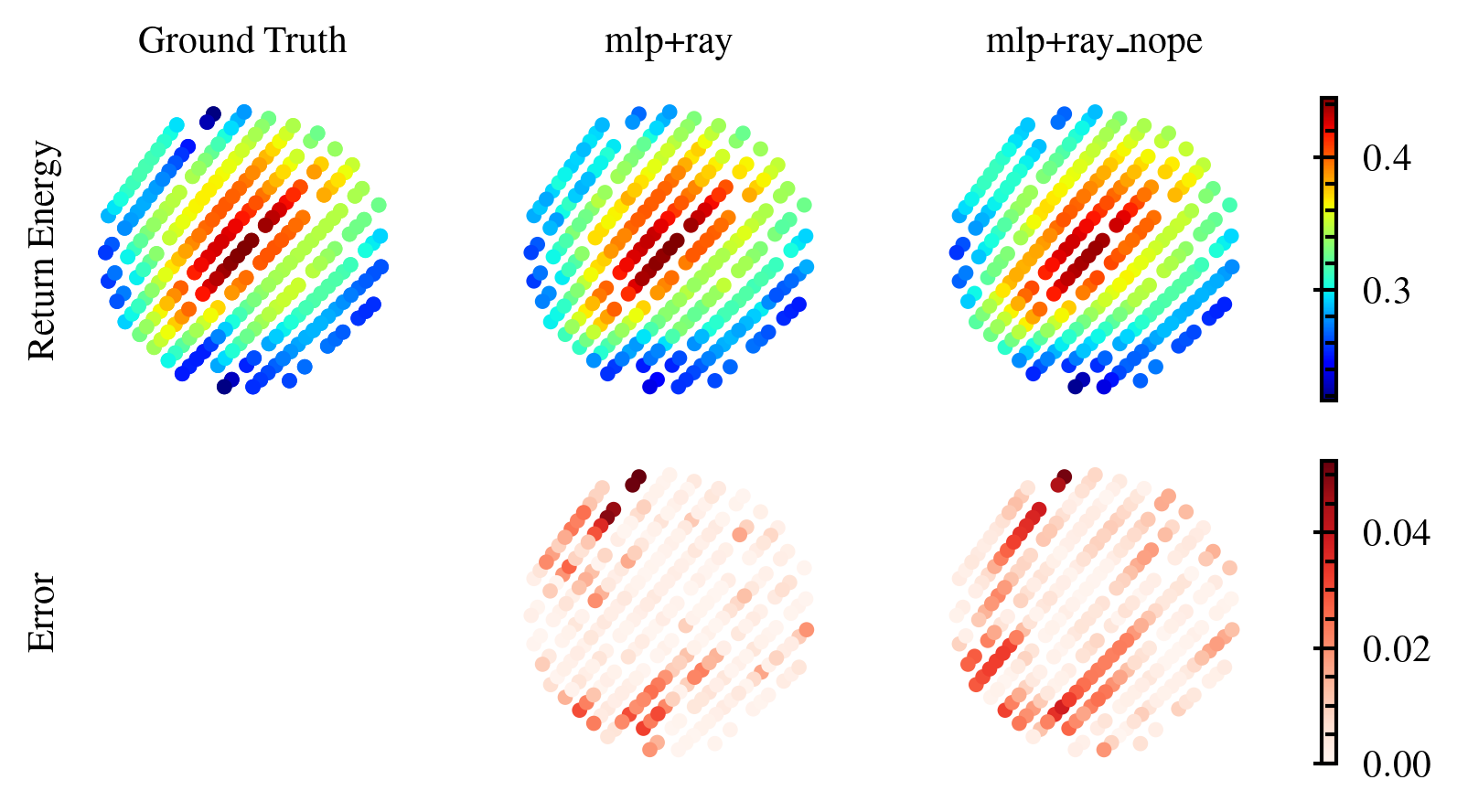}
    \caption{Regression results from MLP}
    \label{fig:regression}
\end{figure}

\begin{figure}[t]
   \centering
    \includegraphics[width=0.5\textwidth]{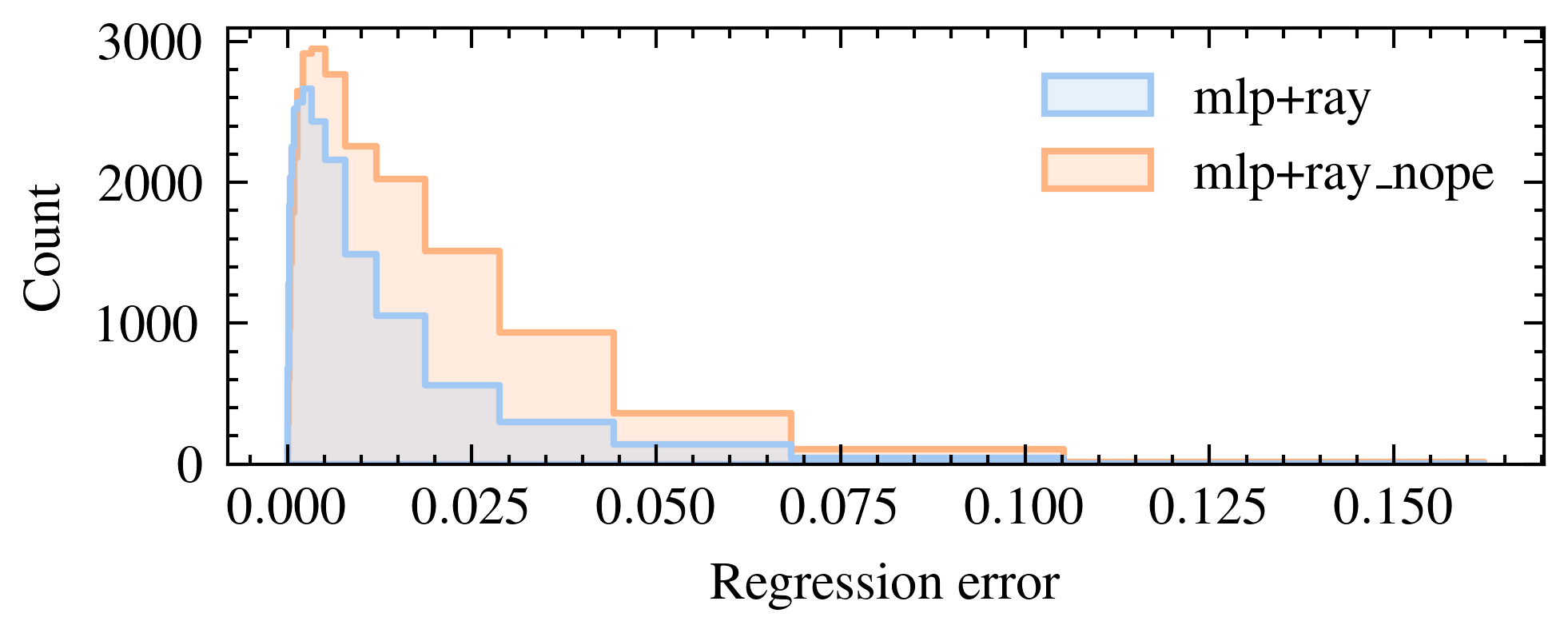}
    \caption{Histogram of regression errors of MLP with/without positional encoding}
    \label{fig:regression_error}
\end{figure}

To further evaluate the quality of the calibration results, we examine the relationship between the target and RADAR ray. The RADAR ray from the sensor origin to the RADAR target center must hit the target and return the energy, as indicated by the high return energy. Fig. \ref{fig:theta} displays the distribution of the LIDAR target centers in the RADAR coordinate after projection using the calibration results. The x-axis represents the radial distance between the LIDAR target center and the RADAR origin. In contrast, the y-axis represents the angle between the RADAR x-y plane and the LIDAR target centers. The two red dashed lines indicate the boundaries where the RADAR ray can hit the target at different distances based on the physical size of the target.
From Fig.\ref{fig:theta}, we observe that the calibration results that only use the reprojection error loss have many points outside the boundary, which is not physically correct. On the other hand, the calibration results of \emph{mlp+ray} and \emph{rep+ray} have most of the projected points within the boundaries. The \emph{mlp} also has most of the points within or close to the boundary, which implies that the objective function is effective.
\begin{figure}[t]
   \centering
    \includegraphics[width=0.5\textwidth]{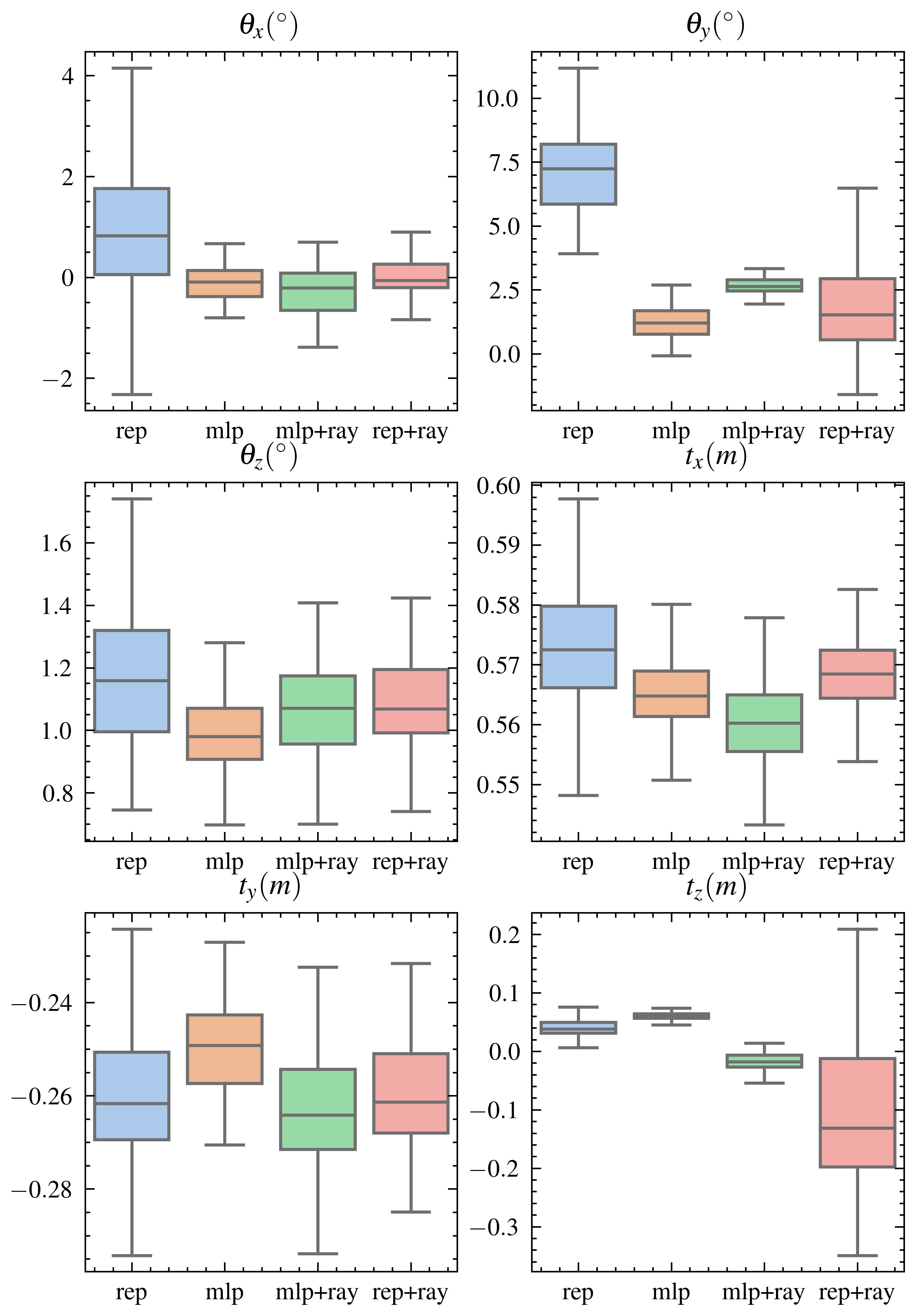}
    \caption{Monte Carlo analysis results for RADAR–LIDAR calibration using different loss configurations}
    \label{fig:mca}
\end{figure}
In order to evaluate the robustness of our method, we conducted a Monte Carlo analysis by randomly subsampling our dataset to 50\% of its original size and ran 100 iterations of optimization on different subsampled datasets. The results are presented as boxplots in Fig. \ref{fig:mca}. As predicted and in line with previous studies \cite{persic_extrinsic_2017,persic_extrinsic_2019}, the parameters $t_x$, $t_y$, and $\theta_z$ that are well-represented by the RADAR measurements exhibit lower variance compared to the other parameters. In contrast, the parameters $t_z$, $\theta_x$, and $\theta_y$ show larger variance. The results from the \emph{rep+ray} experiment display the highest variance in $z$, indicating that the ray-pass loss function is highly sensitive to the amount of data. By incorporating the regression loss, which utilizes both reprojection error and the relationship between the return energy and the position of the target center's surroundings, the variance is reduced compared to only using reprojection error and ray-pass, even with fewer data points.
Finally, Fig.\ref{fig:regression_error} demonstrates that the use of positional encoding results in smaller errors in regression. Several examples of the regression are illustrated in Fig.\ref{fig:regression}.

\section{Conclusion}
In this study, we presented a novel extrinsic 6-DoF calibration method for a RADAR-LIDAR system. Our method used a specially designed calibration target that allowed both sensors to accurately detect and locate the target within their respective operating parameters. The calibration process involved three optimization objectives: reprojection error, regression error, and ray-pass loss. The proposed method was implemented using a popular deep-learning framework and optimized via gradient descent. The experiments conducted on real-world data validated the effectiveness of the proposed method and showed significant improvements in the estimation of extrinsic calibration parameters.

\bibliographystyle{IEEEtran}
\bibliography{IEEEabrv,lidar_radar_calib}

\end{document}